\documentclass[letterpaper]{article}
\usepackage[preprint]{aaai2027}
\usepackage[hyphens]{url}
\usepackage{graphicx}
\usepackage{natbib}
\usepackage{caption}
\usepackage{algorithm}
\usepackage{algorithmic}
\usepackage{newfloat}
\usepackage{listings}
\DeclareCaptionStyle{ruled}{labelfont=normalfont,labelsep=colon,strut=off}
\floatstyle{ruled}
\newfloat{listing}{tb}{lst}{}
\floatname{listing}{Listing}

\usepackage{booktabs}
\usepackage{multirow}
\usepackage{xspace}
\usepackage{amsmath}
\usepackage{amssymb}

\newcommand{\approach}{TRACE\xspace}

\title{\approach: \underline{T}ool-\underline{R}elevance-\underline{A}ware \underline{C}ache \underline{E}viction with Conformal Guarantees for Agentic Inference}
\author{
    Qi Luo\textsuperscript{\rm 1},
    Kunlin Li\textsuperscript{\rm 1},
    Ziwen Wang\textsuperscript{\rm 1},
    Dongsheng Wang\textsuperscript{\rm 2},
    Yun Chen\textsuperscript{\rm 1}\corresponding
}
\affiliations{
    \textsuperscript{\rm 1}The Hong Kong University of Science and Technology (Guangzhou)\\
    \textsuperscript{\rm 2}Tsinghua University\\
    \{qluo615, kli082, zwang578\}@connect.hkust-gz.edu.cn, wds@tsinghua.edu.cn, yunchen@hkust-gz.edu.cn
}

\begin{document}

\maketitle

\begin{abstract}
Agentic large language models equipped with many tools spend most of their KV cache on tool schemas, and this schema-dominated portion sets the practical compression ceiling. Existing compression methods are commonly tuned against cheap proxies such as perplexity, and how much of the schema cache a turn actually needs to preserve task success remains unknown. On the Berkeley Function Calling benchmark, a turn exposes about 28 tools while its reference calls invoke an average of 1.7. The schema cache of the uninvoked tools occupies 77\% to 86\% of the prefix, and removing it at prefill leaves task success essentially unchanged, whereas same-count random removal is catastrophic. We turn this finding into \textbf{\approach}: a lightweight predictor ranks the tools of the current turn, a keep rule prefills the turn's cache with only their schemas, and a conformal layer calibrates the keep size against a risk budget. To our knowledge, this calibration gives the first distribution-free guarantee on the task success an agent loses to cache compression. The best-effort point retains 17\% of the base prefix and 25\% of the long-context prefix at a 3 to 5 point cost, and at risk budget $\alpha=0.10$ the certificate selects a compressed member of the deployed keep-size family in all four model--context settings, a 3.9--6.2$\times$ eviction-only reduction. Per-turn tool relevance, not removed volume, is what makes aggressive schema compression safe, and calibration is what makes it certifiable.

\end{abstract}

\section{Introduction}

Agentic large language models complete real-world multi-step tasks by calling external tools such as web search, code execution, file operations, and messaging APIs \citep{schick2023toolformer,yao2023react,patil2024gorilla,shinn2023reflexion,liu2024agentbench}. To call a tool the model is shown the tool schema, a structured description of its name and parameters. Real agents expose many tools at once to cover complex tasks, on the order of a hundred definitions and tens of thousands of schema tokens \citep{qin2024toolllm,sadani2026toolattention}, and these schemas sit in the KV cache for the whole session and dominate its memory.

KV cache compression is the standard way to cut this cost, by storing the cache at low precision \citep{liu2024kivi,hooper2024kvquant,kang2024gear} or by evicting the cache of low-value tokens \citep{zhang2023h2o,li2024snapkv,xiao2024streamingllm,liu2023scissorhands,ge2024fastgen,cai2024pyramidkv}. But two limitations carry over to agents. First, these methods are tuned and reported against cheap proxies such as perplexity, and compression can silently break instruction following even when such aggregate metrics look benign \citep{chen2025pitfalls}. Second, they treat the prefix as a flat token sequence and do not ask which part of it the current step actually needs. Concurrent work does target tool-schema overhead by compressing schema text or retrieving which tools enter the prompt \citep{sakizli2026toolschema,sadani2026toolattention,lumer2025tooltoagent}, but leaves open the cache-side question of how much of the schema cache a single turn actually needs and what keeping less of it costs in real task success.

\begin{figure}[!t]
\centering
\includegraphics[width=\columnwidth]{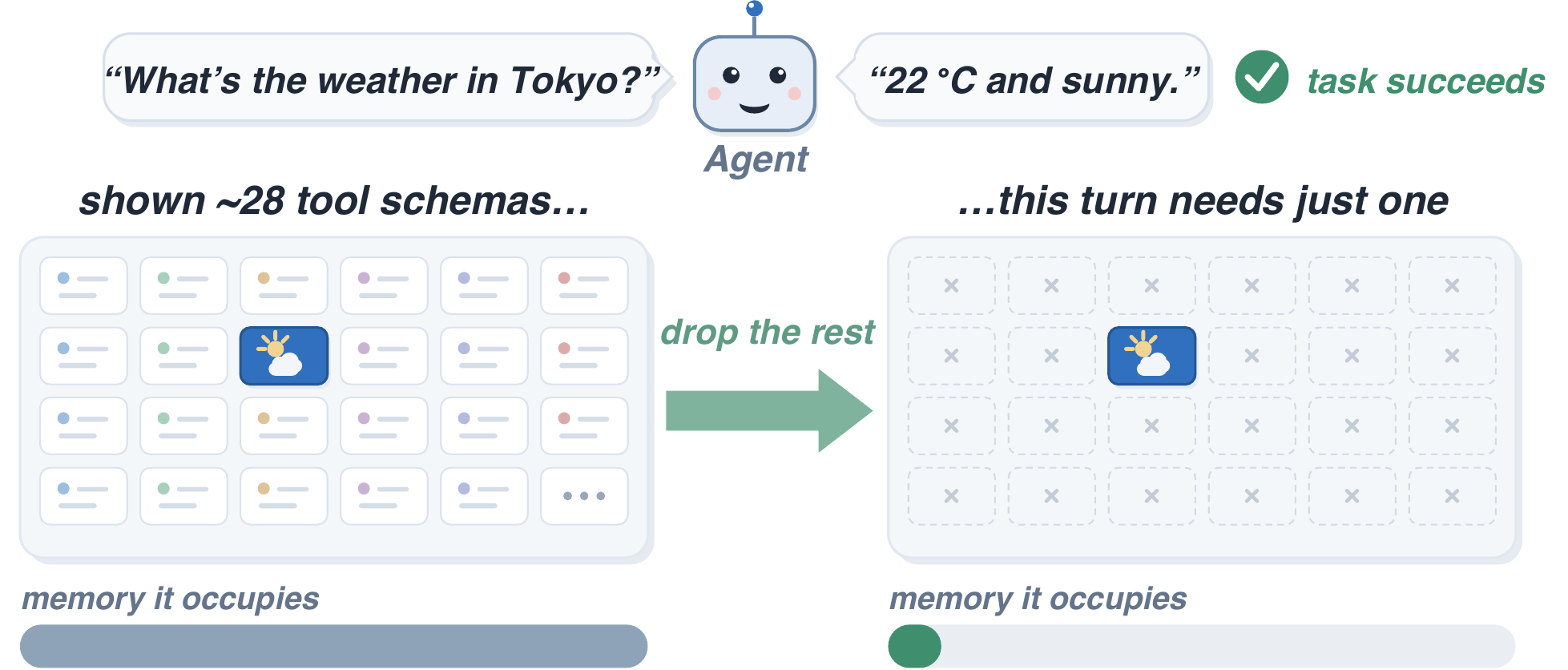}
\caption{Schema headroom per BFCL turn: about 28 available tools versus one typically invoked by the reference calls.}
\label{fig:teaser}
\end{figure}

We study this question on the Berkeley Function Calling benchmark (BFCL) \citep{bfcl2025}, scoring real per-turn task success with the benchmark state checker. Tool schema occupies 92\% of the prefix on the base split and 79\% on the long-context split. Compressing tool results alone preserves success but saves little memory, whereas task success collapses under policies that reach the schema-dominated remainder. This remainder is the practical ceiling we call the tool-schema wall. Yet such cheap proxies miss this constraint.

As shown in Figure~\ref{fig:teaser}, the wall is mostly unused within any single turn. A weather query invokes only the weather tool, and dropping the schemas of the other exposed tools frees most of the schema memory. Per-turn tool usage is highly concentrated. About 28 tools are presented per turn, while the BFCL reference calls invoke only about 1.7 tools on average. To measure the headroom this concentration creates, we run a hindsight oracle that reads each turn's reference calls. It keeps the schema cache of only the invoked tools and evicts the rest, removing 77\% to 86\% of the prefix. Rather than hurting, this eviction nominally improves task-success point estimates by 1.3 to 3.2 points across Qwen2.5 at 7B and 14B. By contrast, evicting the same number of tools at random is catastrophic. Thus, the reference-invoked schemas are sufficient to preserve measured average task success on the evaluated workloads. However, the oracle reads each turn's reference calls, which a deployed agent never sees. A practical policy must instead predict which tools a turn needs. A wrong prediction evicts the schema of a tool the model must call, which is exactly the failure that makes random eviction catastrophic. Safe eviction therefore requires a guarantee that the induced task-success loss stays within a budget.

We turn this finding into \textbf{\approach}, a tool-relevance-aware cache eviction method with three components: (1) a lightweight predictor scores the tools of the current turn, (2) a keep rule retains the schema cache of the top-ranked tools and evicts the rest, and (3) conformal risk control (CRC) \citep{angelopoulos2024conformal} calibrates the keep size offline so that the compression-induced task-success loss stays below a chosen risk budget.

In summary, we make the following contributions:
\begin{itemize}
\item On the BFCL benchmark, we identify the schema-dominated prefix, 79\% to 92\% tool schema, as the tool-schema wall, the practical ceiling on safe agentic KV compression that cheap proxies miss. The wall is mostly unused. A hindsight oracle evicts 77\% to 86\% of the prefix with task success essentially unchanged, whereas same-count random eviction is catastrophic.
\item We propose \approach, which predicts the tools a turn needs, prefills the cache with only their schemas, and calibrates the keep size offline with CRC. To our knowledge, this yields the first distribution-free guarantee on the task success an agent loses to cache compression.
\item The best-effort point of \approach retains 17\% of the base prefix and 25\% of the long-context prefix at a 3 to 5 point cost, far below our attention-score controls at matched memory. At $\alpha=0.10$ the certificate selects a compressed member of the deployed keep-size family in all four model--context settings, a 3.9--6.2$\times$ eviction-only reduction.
\end{itemize}

\section{Related Work}

\paragraph{KV cache compression} There are two mainstream ways to reduce cache cost. Quantization stores the cache at low precision \citep{liu2024kivi,hooper2024kvquant,kang2024gear}. Eviction drops the cache of low-value tokens under attention-based scores \citep{zhang2023h2o,li2024snapkv} or fixed positional rules \citep{xiao2024streamingllm}, while sparse-attention schemes keep the cache but read it selectively \citep{jiang2024minference,tang2024quest}. None of these encodes tool identity, and in our harness the H2O-style and SnapKV-style scores lose substantially more task success than relevance-targeted schema eviction at matched memory. Recent work targets agentic caches specifically, dropping stale tool outputs or retrieved evidence \citep{observationmasking2025,kariyappa2026sidequest}, pruning history tokens by cross-turn intent \citep{li2026intentkv}, exposing primitives for policy-directed cache editing \citep{ma2026leyline}, assigning role-aware precision across the trajectory \citep{triaxialkv}, or managing context segments by lifecycle \citep{xu2026tokenpilot}. These act on non-schema tokens or retain every schema, and none conditions on the per-turn relevance of the schema itself or bounds the resulting task-success loss. Our method is orthogonal to quantization and to serving-level reuse of shared prefixes \citep{kwon2023pagedattention,zheng2024sglang,gim2024promptcache}. Prompt compression \citep{jiang2023llmlingua} and retrieval-augmented context compression \citep{xu2024recomp} shorten the text before it enters the model, scored by perplexity or by a summarizer trained for downstream accuracy, typically without a bound on task-level loss. \approach instead evicts entire tool schemas from a KV cache already built for a structured input, and calibrates the resulting task-success loss with a distribution-free guarantee rather than a heuristic compression ratio.

\paragraph{Tool use and tool retrieval} Agents call external tools \citep{schick2023toolformer,yao2023react}, typically declared to the model as in-context schemas \citep{patil2024gorilla}. When the tool set is large, retrieval selects a subset to place in the prompt, and planning agents decouple which tools to call from the calls themselves \citep{qin2024toolllm,xu2023rewoo}. Concurrent work also compresses the schema text itself \citep{sakizli2026toolschema} or retrieves and gates schemas per turn \citep{sadani2026toolattention,lumer2025tooltoagent}. Prompt-side retrieval is therefore the closest adjacent mechanism. The distinction lies in where the compression acts and in the evaluation target. We measure the cache footprint and the real state-based task loss of per-turn schema retention, realize the policy as compacted prefill without online editing of an existing cache, and calibrate that loss rather than tool recall or exact match. Among tool-use benchmarks \citep{li2023apibank,liu2024agentbench}, we adopt the Berkeley Function Calling benchmark \citep{bfcl2025} for its state checker, which measures real task success rather than a proxy.

\paragraph{Conformal risk control} Conformal prediction gives distribution-free guarantees on prediction sets, and conformal risk control extends this to bound the expected value of a monotone loss \citep{angelopoulos2021gentle,angelopoulos2024conformal}. Concurrent work applies CRC to agent trajectories, calibrating when to accept or intervene on a tool-using run \citep{opoku2026toolchaincrc}, but we are not aware of prior work that applies it to cache compression. Doing so is not a drop-in use of the bound, because the measured task-success loss is neither exchangeable across a task's decision points nor monotone in the keep size. We therefore calibrate over task instances on a pointwise monotone majorant, and the distribution-free guarantee attaches directly to the parameter that controls compression.

\section{The Tool-Schema Wall}
\label{sec:wall}

\subsection{A Motivating Example}

\begin{figure}[t]
\centering
\includegraphics[width=\columnwidth]{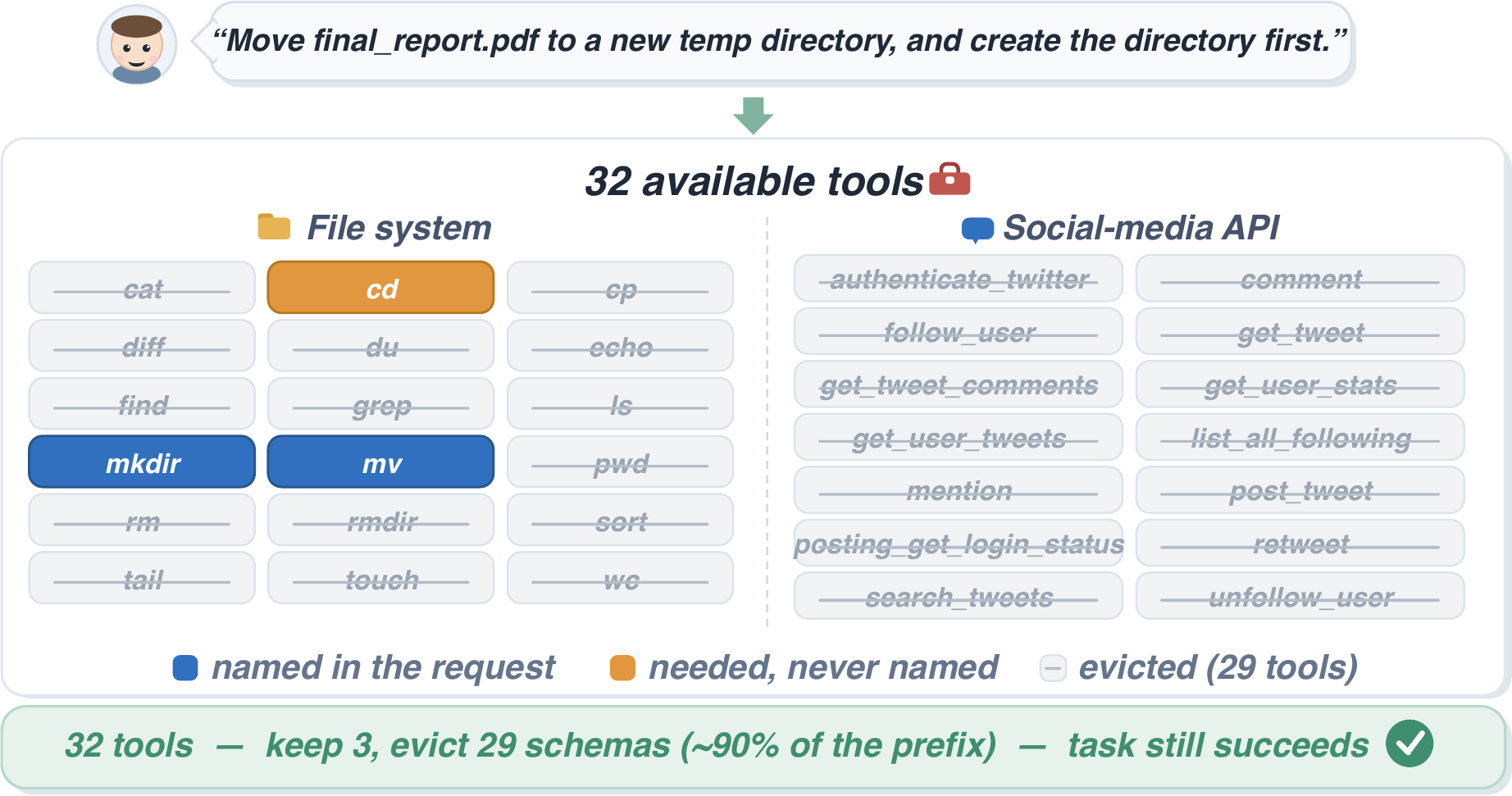}
\caption{Example BFCL turn with 32 available tools and three reference-invoked tools.}
\label{fig:case}
\end{figure}

Figure~\ref{fig:case} shows the first turn of a BFCL task in which the agent is asked to move a file into a new directory. Although the turn exposes 32 tools, the benchmark reference calls invoke only \texttt{cd}, \texttt{mkdir} and \texttt{mv}, leaving the other 29 tools uninvoked. Evicting them would remove most of the prefix cache, but selecting the keep set is non-trivial, because the required navigation call \texttt{cd} is never named in the request. Each of the 29 idle schemas spells out a function name, a natural-language description and a typed parameter list, so even one unused tool adds a non-trivial share of KV-cache tokens before the model ever considers calling it. Because the resident KV footprint scales with every token retained in the prefix, this idle schema mass occupies cache memory for the entire trajectory, not just the turn on which it happens to appear, so the same waste persists across turns even when only a handful of tools stays relevant throughout. This example turn captures both the compression headroom and the prediction problem behind \approach.

\subsection{Measuring Real Task Success under Compression}
\label{subsec:measure}

We study multi-turn tool use on the base and long-context splits of BFCL. Each instance equips the agent with tools from several stateful backend classes, such as a file system, messaging APIs, and a vehicle controller. At each turn the model receives a prefix containing the system message, rendered tool schemas, prior calls and tool results, the current user message and the generation prompt. We call each such generation step a \emph{decision point}. Since eviction will operate on whole tool schemas, we record per-tool token spans inside the schema region of the prefix. Each span is verified by an exact decode-and-retokenize round trip against the original prefix, which passes on every decision point.

We measure real task success with the official BFCL state checker. BFCL provides reference function calls for each turn. At each decision point we replay earlier reference calls to recover the pre-turn state, execute the model calls and current-turn reference calls in isolated backends, and compare the resulting states and responses. A turn succeeds when both match, allowing calls that differ in form but reach the same state.

Absolute success cannot be the target of a compression guarantee. Even with a near-lossless 4-bit cache \citep{liu2024kivi}, the Qwen2.5-7B agent succeeds on only about 47\% of base turns, reflecting baseline capability rather than compression damage. We therefore derive two quantities from per-turn success, a signed regret that we report and a non-negative loss that the guarantee bounds. Over $N$ decision points, let $b_i\in\{0,1\}$ mark whether the model with the uncompressed cache succeeds at decision point $i$ and $c_i(\pi)\in\{0,1\}$ whether it succeeds under a compression policy $\pi$. The task-success regret of $\pi$ is
\begin{equation}
\mathrm{Regret}(\pi)=\frac{1}{N}\sum_{i=1}^{N}\big(b_i-c_i(\pi)\big),
\label{eq:regret}
\end{equation}
which is signed and can be slightly negative when removing irrelevant schema reduces distraction. The conformal loss of $\pi$ is
\begin{equation}
L(\pi)=\frac{\sum_{i=1}^{N} b_i\,\big(1-c_i(\pi)\big)}{\sum_{i=1}^{N} b_i},
\label{eq:closs}
\end{equation}
the fraction of the baseline-correct decision points that $\pi$ breaks.
We use this quantity to describe frontiers. For the formal CRC guarantee, Section~\ref{sec:approach-crc} first averages these failures within each task instance so that correlated turns do not enter calibration as independent samples.

\subsection{Schema Sets the Compression Ceiling}

\begin{table}[t]
\centering
\footnotesize
\setlength{\tabcolsep}{2pt}
\begin{tabular}{@{}lcccc@{}}
\toprule
\multirow{2}{*}{\textbf{quantization policy}} & \multicolumn{2}{c}{\textbf{base}} & \multicolumn{2}{c}{\textbf{long context}} \\
\cmidrule(lr){2-3}\cmidrule(lr){4-5}
& \textbf{ratio} & \textbf{success} & \textbf{ratio} & \textbf{success} \\
\midrule
all KV at 4 bit & 4.0$\times$ & 47.2\% & 4.0$\times$ & 39.4\% \\
tool-result KV at 2 bit & 4.0$\times$ & 46.9\% & 4.3$\times$ & 39.7\% \\
half of prefix KV at 2 bit & 5.3$\times$ & 16.1\% & 5.3$\times$ & 18.7\% \\
all KV at 2 bit & 8.0$\times$ & 0.3\% & 8.0$\times$ & 0.7\% \\
\bottomrule
\end{tabular}
\caption{Quantization policies on Qwen2.5-7B.}
\label{tab:schema-wall}
\end{table}

On both splits the cache is dominated by tool schema, about 92\% of the prefix on the base split and 79\% on the long-context split, while tool results, the largest non-schema segment, take only 1\% and 17\%, respectively. As shown in Table~\ref{tab:schema-wall}, mixed-precision quantization probes \citep{liu2024kivi} make the bottleneck concrete. Moving tool results alone to 2 bit preserves task success but reaches only 4.0$\times$ and 4.3$\times$ over 16-bit KV. Quantizing half of the prefix KV reaches 5.3$\times$ but collapses task success to 16.1\% and 18.7\%, and quantizing all KV to 2 bit lowers it to 0.3\% and 0.7\%. The quantized half is 91\% schema on base and 72\% on lc, so the intervention is not schema-only and we attribute the collapse to the schema-dominated remainder rather than to schema KV alone. For the evaluated policies and workloads, useful non-schema compression is exhausted before schema can be touched. We call this empirical ceiling the \emph{tool-schema wall}.

\begin{table}[ht]
\centering
\small
\setlength{\tabcolsep}{3pt}
\begin{tabular}{@{}clccc@{}}
\toprule
$\boldsymbol{\alpha}$ & \textbf{metric} & \textbf{ratio} & \textbf{legality loss} & \textbf{task loss} \\
\midrule
0.10 & legality     & 4.4$\times$ & 5.8\%  & 12.6\% \\
0.10 & task success & 4.0$\times$ & 0.0\%  & 0.0\% \\
0.20 & legality     & 5.0$\times$ & 14.1\% & 30.8\% \\
0.20 & task success & 4.4$\times$ & 5.8\%  & 12.6\% \\
\bottomrule
\end{tabular}
\caption{Policy selection under legality and task-success metrics on Qwen2.5-7B base. Losses are relative to the all-4-bit reference.}
\label{tab:proxy}
\end{table}

Cheap proxies obscure this ceiling. Table~\ref{tab:proxy} picks an operating point on the same mixed-precision grid with the calibrated selection of Section~\ref{sec:approach-crc}, scored by either output legality or BFCL task success. Losses count new failures among decision points where the all-4-bit reference satisfies the corresponding metric. At risk budget $\alpha=0.10$, legality selects 4.4$\times$, whose 12.6\% task loss exceeds the budget, while task-success selection stays at the 4.0$\times$ reference. The mismatch recurs at $\alpha=0.20$ and on 14B, where a legality-selected 4.4$\times$ policy incurs 15.5\% task loss.

\subsection{The Wall Is Mostly Unused}

Uniform quantization damages the schema of every tool, invoked or not, whereas eviction can be selective. A turn exposes between 18 and 39 tools across the benchmark, about 28 on average, while the BFCL reference calls invoke only 1.7 tools on average, with a median of 1 and at most 7. At decision point $i$, let $U_i$ be the set of tool names appearing in that turn's BFCL reference calls. The hindsight oracle keeps the schema spans of tools in $U_i$ together with the non-schema prefix and evicts all other tool-schema spans, removing 77\% to 86\% of the prefix depending on the split. We call it the \emph{oracle} for brevity, though reference calls need not be the unique valid action sequence.

Oracle eviction changes the Qwen point estimates by only a few points, while evicting the same number of tools at random is catastrophic, so it is relevance, not removed volume, that determines the damage. A predictor need not be minimal either. Keeping one extra random tool also stays within noise on 7B, so missed tools are the risk and extras are nearly free. A predictor thus only needs to place the reference-invoked tools inside a keep set that may hold a few extras. This headroom is what \approach captures. Predicting implicit tools such as the \texttt{cd} of the opening example is the residual difficulty, quantified in Section~\ref{subsec:analysis}.

\section{\approach}

\begin{figure*}[t]
\centering
\includegraphics[width=\textwidth]{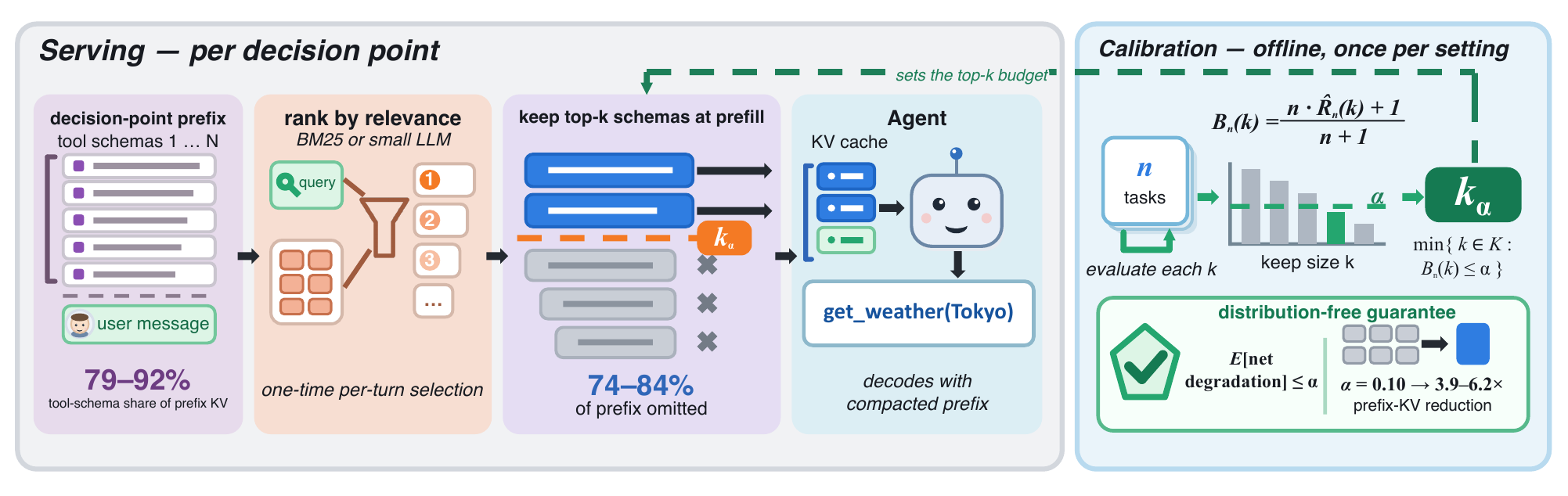}
\caption{Overview of \approach. At serving time, a lightweight predictor ranks the tools of the current turn and the keep rule prefills the cache with only the top-$k$ schemas. An offline conformal calibration selects the keep size $k_\alpha$ that keeps the compression-induced task-success loss within the risk budget $\alpha$.}
\label{fig:arch}
\end{figure*}

As shown in Figure~\ref{fig:arch}, \approach has three components: (1) a relevance predictor that scores the tools of the current turn, (2) a keep rule that retains the schema cache of the top-ranked tools and evicts the rest, and (3) a conformal layer that calibrates the keep size offline against a risk budget.

\subsection{Relevance Predictor and Keep Rule}

The predictor reads the current user text and the tool catalog and ranks tools by relevance. We study two rankers: (1) a lexical BM25 score over tool name and description \citep{robertson2009bm25}, which provides a dense keep-size grid, and (2) a small instruction model, Qwen2.5-1.5B-Instruct \citep{qwen2025}, prompted in a \emph{select} mode that lists only the tools it judges necessary or a \emph{full-ranking} mode that requests a complete order, with both prompts listed in Appendix~A.7. The full-ranking mode often emits only the top of the order and is evaluated best-effort. The select mode is our deployed predictor, and its keep sizes form the nested family that the conformal layer calibrates. Given a ranking, the keep rule keeps the schema of the top $k$ tools and evicts the rest, so $k$ is the parameter that controls the compression rate. The policy is realized at prefill, building the cache for each turn from a compacted prefix that contains only the kept schemas. Our harness uses an attention mask verified equivalent to compacted prefill (Appendix~A.8). Editing an already-populated cache is a distinct mechanism, since retained KV encodes attention into the removed spans, and is not evaluated here.

\subsection{Conformal Calibration}
\label{sec:approach-crc}

CRC requires exchangeable units with a loss that is monotone in the policy parameter \citep{angelopoulos2024conformal}. Decision points from one multi-turn task are correlated, and individual measured loss curves can contain decoding-induced non-monotone reversals. We address both conditions by calibrating over task instances and a pointwise monotone majorant.

For task instance $j$, let $\mathcal{C}_j$ be its decision points that the model with the uncompressed cache gets right. We call the instance eligible when $\mathcal{C}_j$ is nonempty, and define its compression-induced loss as
\begin{equation}
\ell_j(k)=\frac{1}{|\mathcal{C}_j|}\sum_{i\in\mathcal{C}_j}\bigl(1-c_i(k)\bigr)\in[0,1].
\label{eq:instance-loss}
\end{equation}
This is the instance-weighted analogue of Eq.~\ref{eq:closs}. Because increasing $k$ need not lower the measured loss on every decoded sample, we calibrate on the pointwise majorant $\widetilde{\ell}_j(k)=\max_{k'\ge k}\ell_j(k')$, which is non-increasing in $k$ and upper-bounds the measured loss. The candidate grid includes a keep-all endpoint with $\ell_j=0$, so an unsupported risk budget produces a conservative fallback rather than an uncertified compressed point.

Let $\hat{R}_n(k)=n^{-1}\sum_{j=1}^{n}\widetilde{\ell}_j(k)$ on $n$ exchangeable calibration instances. We pick the smallest keep size
\begin{equation}
k_\alpha=\min\Big\{\,k:\;\frac{n}{n+1}\,\hat{R}_n(k)+\frac{1}{n+1}\le\alpha\,\Big\},
\label{eq:crc-select}
\end{equation}
the most aggressive eviction whose finite-sample-corrected calibration risk meets $\alpha$. Applying CRC to $\widetilde{\ell}$ controls the raw loss on a fresh eligible task instance,
\begin{equation}
\mathbb{E}\big[\ell_{n+1}(k_\alpha)\big]
\le \mathbb{E}\big[\widetilde{\ell}_{n+1}(k_\alpha)\big]\le\alpha .
\label{eq:crc-bound}
\end{equation}
The expectation is marginal over the calibration sample and a new instance drawn from the same exchangeable distribution. It does not require every finite held-out split to fall below $\alpha$. The bound concerns compression-induced task-success loss rather than tool recall, and it remains valid for a weak predictor because the selected policy can increase $k$ or keep all schemas.

\begin{algorithm}[!t]
\caption{\textbf{\approach{} Calibration and Serving}}
\label{alg:main}
\begin{algorithmic}[1]
\REQUIRE predictor $r$, risk budget $\alpha$, nested keep-size family $K$ with a keep-all endpoint, $n$ calibration task instances
\ENSURE keep size $k_\alpha$ with $\mathbb{E}[N_{n+1}(k_\alpha)]\le\alpha$
\STATE \textit{Calibration (offline, once)}
\FOR{each calibration instance $j$}
  \STATE for each $i\in\mathcal{T}_j$ and $k\in K$: keep top-$k$ by $r$, regenerate, score $c_i(k)$
  \STATE $N_j(k)\gets\max\big(0,\,|\mathcal{T}_j|^{-1}\sum_{i\in\mathcal{T}_j}(b_i-c_i(k))\big)$,\quad $\widetilde N_j(k)\gets\max_{k'\ge k}N_j(k')$ \COMMENT{Eq.~\ref{eq:netpos}}
\ENDFOR
\STATE $\hat{R}_n(k)\gets\frac{1}{n}\sum_{j=1}^{n}\widetilde N_j(k)$ \COMMENT{keep-all endpoint has $N_j=0$}
\STATE $k_\alpha\gets\min\{\,k\in K:\frac{n}{n+1}\hat{R}_n(k)+\frac{1}{n+1}\le\alpha\,\}$ \COMMENT{Eq.~\ref{eq:crc-select}}
\STATE \textit{Serving (per decision point)}
\STATE keep the top-$k_\alpha$ tools by $r$, evict the other schema spans, decode from the reduced cache
\end{algorithmic}
\end{algorithm}

\paragraph{Certifying the deployed policy} Eq.~\ref{eq:instance-loss} charges every broken baseline-correct turn, including breaks that decoding jitter offsets with repairs elsewhere in the same instance. We therefore also calibrate the \emph{net-degradation} loss
\begin{equation}
N_j(\pi)=\max\Big(0,\ \tfrac{1}{|\mathcal{T}_j|}\textstyle\sum_{i\in\mathcal{T}_j}\big(b_i-c_i(\pi)\big)\Big),
\label{eq:netpos}
\end{equation}
the positive part of the signed regret of instance $j$ over all its turns $\mathcal{T}_j$, which is defined for every instance and vanishes at keep-all. Because Eq.~\ref{eq:crc-select} needs only a nested family, it applies directly to the deployed keep-size family, $\{$select-$k3\subseteq k5\subseteq k8$, keep-all$\}$. The select mode and its prompt are fixed design choices and the keep size is chosen inside the calibration, so point selection is covered by the finite-sample correction, in the spirit of learn-then-test \citep{angelopoulos2021learn}. The selector-family calibration under this loss is our primary certificate (Section~\ref{subsec:crc-exp}), and the stricter breakage-loss calibration of the BM25 grid is reported in Appendix~A.4. Algorithm~\ref{alg:main} summarizes the deployed calibration and serving.

\section{Experiments}
\label{sec:experiments}

\subsection{Setup}
\label{subsec:exp-setup}

\paragraph{Models and workloads} We evaluate Qwen2.5-Instruct \citep{qwen2025} at 7B and 14B on BFCL v4 \citep{bfcl2025}, release 2026.3.23, with the official harness. The base and long-context (lc) splits contain 386 and 329 decision points. All lc generation omits the 19 prefixes over 30k tokens, which exceed the single-GPU memory of our regeneration harness, leaving 310 decision points. Only the prefix-composition statistics use the full split. We also replicate the causal test on 330 \texttt{multi\_turn\_miss\_param} decision points and on Llama-3.1-8B-Instruct \citep{llama3herd2024} first turns.

\paragraph{Metrics} We regenerate each turn with greedy decoding, score the calls with the BFCL state checker, and report the signed regret of Eq.~\ref{eq:regret}. Risk-controlled results report the calibrated losses of Section~\ref{sec:approach-crc} at risk budget $\alpha$, and memory as the retained prefix fraction projected from compacted token counts.

\paragraph{Baselines} We compare \approach with the hindsight oracle of Section~\ref{sec:wall}, simulated fake-quant low-bit KV storage \citep{liu2024kivi}, and H2O-style \citep{zhang2023h2o} and SnapKV-style \citep{li2024snapkv} attention eviction at token and tool levels, implemented in our masked-generation harness based on the original papers (Appendix~A.5). Their quadratic attention maps limit comparisons to 382 base decision points with prefixes below 8k tokens, and their eager backend shifts the level, reaching a 40.1\% no-op baseline versus 49.2\% under the default kernel. Every regret is measured against the uncompressed baseline of its own backend, and matched-memory comparisons across the two families are approximate.

\paragraph{Implementation} Experiments run on single 80GB A800 GPUs under Linux, with bfloat16 agent weights, PyTorch 2.6, CUDA 12.4, and Hugging Face Transformers 5.12. The 14B lc generation uses a single H100 for faster compute. A no-op mask matches uncompressed success, and masked eviction is validated against physical prefix compaction. Our harness realizes serving-time eviction as compacted prefill rather than by modifying an existing cache. HBM is measured for the oracle keep set, and practical-policy memory and 4-bit ratios are projections.

\subsection{Tool Relevance Enables Safe Schema Eviction}
\label{subsec:free}

\begin{table}[t]
\centering
\small
\setlength{\tabcolsep}{5pt}
\begin{tabular}{lcccc}
\toprule
\textbf{Model} & \textbf{success} & \textbf{oracle} & \textbf{+1 rand} & \textbf{random} \\
\midrule
Qwen2.5-7B base   & 49.2\% & $-1.3$ & $-1.8$ & $+43.0$ \\
Qwen2.5-7B lc     & 40.3\% & $-3.2$ & $-2.6$ & $+35.5$ \\
Qwen2.5-14B base  & 49.5\% & $-1.3$ & $-1.0$ & $+43.5$ \\
Qwen2.5-14B lc    & 43.2\% & $-1.3$ & $-1.3$ & $+40.6$ \\
\bottomrule
\end{tabular}
\caption{Qwen uncompressed success and hindsight-oracle/control regret (pp) over all decision points.}
\label{tab:app-eviction}
\end{table}

Table~\ref{tab:app-eviction} compares the hindsight oracle with two controls. The +1 rand control keeps the oracle set plus one extra random tool, and the random control keeps a randomly chosen set of the same size as the oracle keep set and evicts the rest. The oracle removes 77\%--86\% of the prefix at point-estimate regret between $-3.2$ and $-1.3$ across Qwen scales and contexts, while same-count random eviction loses 35.5--43.5 points. Negative regret means task success under eviction is nominally higher than with the full cache. Which schemas are evicted matters more than eviction volume, and the near-zero oracle estimates indicate headroom rather than an improvement from eviction. Task-instance intervals for 7B are in Appendix~A.2, and the 14B rows are point-estimate replications. The pattern also holds beyond Qwen and beyond the two main splits. On Llama-3.1-8B first turns, oracle eviction yields $-5.0$/$-7.0$ points on base/lc while same-count random eviction costs 33.0 on both, and on the \texttt{multi\_turn\_miss\_param} setting the oracle evicts 84.5\% of the prefix at $-2.7$ points versus $+35.8$ under random (Appendix~A.2).

\subsection{Practical Relevance Prediction}
\label{subsec:analysis}

\begin{figure}[t]
\centering
\includegraphics[width=\columnwidth]{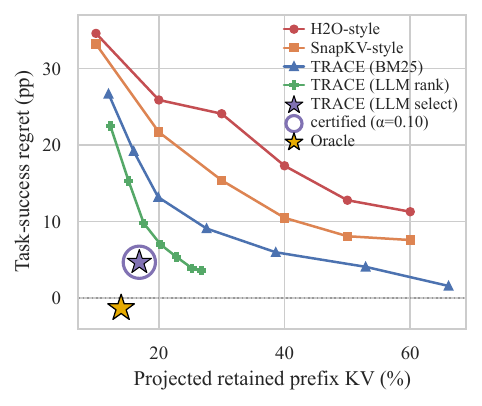}
\caption{Retained prefix KV versus task-success regret (pp, lower is better) on Qwen2.5-7B base. The ring marks the $\alpha{=}0.10$ net-degradation certificate within the deployed keep-size family.}
\label{fig:frontier}
\end{figure}

\paragraph{Predictor frontier} Figure~\ref{fig:frontier} measures how much oracle headroom the practical predictors recover. The small instruction model in full-ranking mode reaches 3.6 points of regret at 27\% retained cache, whereas BM25 still incurs 4.1 points at 53\%. The select mode of the small model gives the strongest best-effort point, 4.7 points of regret at 17\% retained on base and 3.2 points at 25\% on lc, a projected 5.9$\times$ and 3.9$\times$ reduction in prefix-cache footprint. Storing the retained KV at 4 bits would project roughly 24$\times$ versus a 16-bit cache. The frontier replicates toward each oracle ceiling on Qwen2.5-14B and Llama-3.1-8B (Appendix~A.3), with operating points on different evaluation subsets.

\paragraph{Matched-memory attention-score controls} Near 17\%--20\% retained cache, the token-level H2O-style and SnapKV-style scores lose 22--26 points and \approach loses 4.7, each measured against the baseline of its own backend (Section~\ref{subsec:exp-setup}). Even at 60\% retained, the attention scores lose 11 and 8 points. Uniform quantization saturates near 4.4$\times$, so it cannot reach the relevance-targeted frontier. Because token- versus tool-level eviction is a potential confound, Table~\ref{tab:ablation} gives the same attention signals the same schema grouping and keep sizes as \approach.

\begin{table}[ht]
\centering
\small
\begin{tabular}{lcc}
\toprule
\textbf{Method} & \textbf{retained} & \textbf{regret (pp)} \\
\midrule
H2O-style per-tool, $k{=}3$    & 19\% & 33.2 \\
H2O-style per-tool, $k{=}8$    & 36\% & 28.8 \\
\midrule
SnapKV-style per-tool, $k{=}3$ & 19\% & 10.2 \\
SnapKV-style per-tool, $k{=}8$ & 36\% & 6.3 \\
\midrule
\approach (BM25), $k{=}3$ & 20\% & 13.2 \\
\approach (BM25), $k{=}8$ & 39\% & 6.0 \\
\textbf{\approach (LLM select)} & \textbf{17\%} & \textbf{4.7} \\
\bottomrule
\end{tabular}
\caption{Per-tool scoring ablation on Qwen2.5-7B base. Each regret is measured against the baseline of its own backend.}
\label{tab:ablation}
\end{table}

Grouping improves the SnapKV-style variant to 10.2 points of regret at 19\% retained and 6.3 at 36\% while the H2O-style score stays poor, so the advantage has two sources. Whole-schema granularity avoids destructive token fragmentation, and relevance prediction improves the ordering within it.

\paragraph{Where the predictor fails} At $k=5$, all reference-invoked tools appear in the keep set for about 60\% of BM25 decision points and 63\% of small-model decisions (Appendix~A.7). Most misses are implicit procedural tools absent from the request, such as the \texttt{cd} before a file operation. Keeping previously called tools or a small global core set does not close this gap. Predictor recall, rather than a lack of schema headroom, is therefore the main limitation of the practical frontier. Closing it likely needs signal beyond the current turn text, such as a model that infers required procedural tools from the semantics of the requested action, or an explicit schema graph that encodes which tools typically precede which. Either extension keeps the same keep rule and calibration, and only changes the ranking that feeds it. We leave both to future work.

\subsection{Risk-Controlled Policy Selection with CRC}
\label{subsec:crc-exp}

\begin{table}[ht]
\centering
\footnotesize
\setlength{\tabcolsep}{3.2pt}
\begin{tabular}{l ccccc}
\toprule
\textbf{Setting} & $\boldsymbol{k}$ & \textbf{evict} & \textbf{retain} & \textbf{risk} & \textbf{held-out} \\
\midrule
7B base & 3 & 6.2$\times$ & 16.1\% & 0.099 & 0.077 \\
7B lc & 3 & 4.1$\times$ & 24.7\% & 0.090 & 0.081 \\
14B base & 5 & 5.9$\times$ & 16.9\% & 0.097 & 0.071 \\
14B lc & 8 & 3.9$\times$ & 25.8\% & 0.098 & 0.066 \\
\bottomrule
\end{tabular}
\caption{The $\alpha=0.10$ net-degradation certificate in the four model--context settings.}
\label{tab:crc-main}
\end{table}

\paragraph{Primary protocol} CRC treats each complete BFCL task instance as one exchangeable unit, and under the net-degradation loss of Eq.~\ref{eq:netpos} all 140 base and 120 lc instances qualify. Selection applies the pointwise majorant of Section~\ref{sec:approach-crc} to the deployed nested keep-size family with its keep-all endpoint. Table~\ref{tab:crc-main} reports calibration on all instances. At $\alpha=0.10$ a compressed member certifies in all four settings, a 3.9--6.2$\times$ eviction-only reduction that retains 16--26\% of the prefix, with corrected risks 0.090--0.099. Mean net degradation on the held-out 20\% of instances stays at 0.066--0.081 across 400 instance-disjoint 80/20 splits, small calibration sets abstain to keep-all on 9--47\% of splits, and at $\alpha=0.05$ every setting abstains.

\paragraph{The stricter breakage semantics} Eq.~\ref{eq:instance-loss} instead charges every broken baseline-correct turn, even when decoding jitter repairs as many elsewhere in the instance. That loss has a measured floor. The minimal reference keep set itself breaks 7.8\%/6.0\% of baseline-correct turns on base/lc, within 0.014 of the $\alpha=0.10$ threshold, so no eviction policy, not even the oracle, can be certified deeply under it. We therefore report its calibration in Appendix~A.4. The BM25 grid backs off to conservative keep sizes at $\alpha=0.10$, and the small-model ranking certifies 2.8--4.4$\times$ at $\alpha=0.15$--$0.20$.

\subsection{Always-On Rollouts and Selection Overhead}

\paragraph{Compression always on} The main protocol replays reference history to isolate each turn, so we also run 100 base instances, 330 turns, from self-generated calls and backend observations, with eviction active at every generation step, testing $k=16$, which is the isolated-turn $\alpha=0.10$ selection, and $k=8$ as a stress point. Appendix~A.6 reports 1.8 and 5.8 points of turn-level regret, and total calls differ by at most 3\%, so we observe no call-count inflation. Because only six baseline episodes succeed end to end we report turn-level numbers.

\begin{table}[ht]
\centering
\small
\begin{tabular}{lcccc}
\toprule
\textbf{Method} & \textbf{mean} & \textbf{median} & \textbf{p90} & \textbf{max} \\
\midrule
BM25        & 1.2  & 1.2  & 1.6  & 2.0 \\
LLM select  & 227  & 166  & 449  & 975 \\
LLM rank    & 399  & 283  & 963  & 1108 \\
\midrule
Attention scoring & 717 & 749 & 994 & 1079 \\
\bottomrule
\end{tabular}
\caption{Per-turn selection latency in ms.}
\label{tab:app-latency}
\end{table}

\paragraph{Selection overhead} Table~\ref{tab:app-latency} reports one-time per-turn selection cost. Median latency is 1.2 ms for CPU BM25, 166 ms for small-model select mode and 283 ms for full ranking on the A800. In our eager implementation, the global attention-scoring pass shared by the H2O-style and SnapKV-style baselines costs 749 ms on a 96-decision-point subset, about 620$\times$ BM25, and its materialized map does not fit beyond 7B. Official kernels would lower this constant, but \approach structurally needs no attention materialization.

\subsection{Limitations and Future Work}

All quantitative results are on BFCL. The causal premise replicates across three workloads and two model families, and an independent agent benchmark is the immediate next step. The guarantee is marginal over exchangeable task instances from the calibrated distribution, so a new model or workload requires recalibration, and with small calibration sets CRC falls back to keep-all rather than certify an unsupported budget. On the serving side, our harness realizes eviction at generation time through compacted prefill, validated against physical compaction on the oracle keep set. An online cache-editing runtime and low-bit storage are the natural engineering steps toward deployment.

\section{Conclusion}

This paper studied agentic KV cache compression under real task success and located the bottleneck in the schema-dominated prefix. A turn exposes about 28 tools yet invokes an average of 1.7, and a hindsight oracle that keeps only the invoked schemas evicts 77--86\% of the prefix at near-zero cost, while same-count random eviction is catastrophic. \approach converts this headroom into a practical policy. A lightweight relevance predictor selects the schemas a turn needs, and conformal risk control calibrates the keep size. To our knowledge, this gives the first distribution-free guarantee on the task success an agent loses to cache compression. On Qwen2.5-7B the best-effort point retains 17\% of the base prefix and 25\% of the long-context prefix at a 3--5 point cost, far below our attention-score controls at matched memory, and at $\alpha{=}0.10$ the net-degradation certificate selects a compressed member of the deployed keep-size family in all four settings, a 3.9--6.2$\times$ eviction-only reduction, with mean held-out loss under budget across repeated splits. Per-turn relevance, not removed volume, is what makes aggressive schema eviction safe, and calibration is what makes it certifiable. The gap that remains between the predictors and the oracle is concentrated on implicit procedural tools, and closing it with stronger predictors tightens the certified operating point within the same guarantee.

\bibliography{references}

\FloatBarrier
\clearpage
\appendix
\setcounter{table}{0}
\setcounter{figure}{0}
\renewcommand{\thetable}{A\arabic{table}}
\renewcommand{\thefigure}{A\arabic{figure}}
\section{Reproducibility and Additional Details}

This appendix collects the per-condition, per-split and per-model numbers behind the main-text summaries, together with the implementation details needed to reproduce them. All results use the experimental protocol of the main paper.

\subsection{Per-Tool Schema Spans}
\label{app:spans}

We recover per-tool token spans without re-running the rollout. We decode the stored prefix back to its rendered string, re-tokenize it with character offsets and accept a span only when the recovered token IDs exactly match the original prefix. Inside the tool region we anchor each tool on its name string and assign it the surrounding span. This exact round-trip check passes on all decision points for the models we use. The tool region is delimited per family, since the chat template renders the schemas differently, and for a model whose template emits multi-line objects each tool span is cut at object boundaries rather than at name midpoints.

\subsection{Full Eviction Results}

\begin{table}[ht]
\centering
\small
\setlength{\tabcolsep}{1.5pt}
\begin{tabular}{lcc}
\toprule
\textbf{Condition} & \textbf{base} & \textbf{lc} \\
\midrule
\multicolumn{3}{l}{\textit{Qwen2.5-7B, oracle and controls}} \\
oracle  & $-1.3$\,$[-4.4,\,1.8]$ & $-3.2$\,$[-6.8,\,0.3]$ \\
+1 rand & $-1.8$\,$[-4.9,\,1.0]$ & $-2.6$\,$[-6.2,\,1.0]$ \\
random  & $43.0$\,$[38.0,\,48.0]$ & $35.5$\,$[29.8,\,41.4]$ \\
\midrule
\multicolumn{3}{l}{\textit{Llama-3.1-8B first turns}} \\
oracle  & $-5.0$\,$[-9.5,\,-1.0]$ & $-7.0$\,$[-11.5,\,-2.5]$ \\
+1 rand & $-2.5$\,$[-6.5,\,1.5]$ & $-2.0$\,$[-7.0,\,3.0]$ \\
random  & $33.0$\,$[26.5,\,40.0]$ & $33.0$\,$[26.0,\,40.0]$ \\
\midrule
\multicolumn{3}{l}{\textit{\approach operating points, 7B}} \\
LLM select    & $4.7$\,$[1.1,\,8.1]$ & $3.2$\,$[-0.7,\,7.1]$ \\
BM25 $k{=}16$ & $1.6$\,$[-1.3,\,4.2]$ & $1.6$\,$[-1.3,\,4.3]$ \\
BM25 $k{=}8$  & $6.0$\,$[2.1,\,9.6]$ & $5.5$\,$[1.3,\,9.6]$ \\
\midrule
\multicolumn{3}{l}{\textit{Attention controls, 7B base, eager backend}} \\
H2O-style 20\%             & $25.9$\,$[20.8,\,31.1]$ & --- \\
SnapKV-style 20\%          & $21.7$\,$[16.7,\,26.5]$ & --- \\
H2O-style tool $k{=}8$     & $28.8$\,$[23.8,\,34.0]$ & --- \\
SnapKV-style tool $k{=}8$  & $6.3$\,$[2.9,\,10.0]$   & --- \\
\bottomrule
\end{tabular}
\caption{Task-instance cluster-bootstrap 95\% CIs for task-success regret (pp).}
\label{tab:app-ci}
\end{table}

Table~\ref{tab:app-ci} reports task-instance cluster-bootstrap intervals for the oracle conditions, the practical operating points, and the attention-score controls, the latter scored against their own eager-backend baseline. The 7B oracle and +1-random intervals include zero, while the matched-random intervals remain far above zero. Among the practical policies, the certified $k{=}16$ point includes zero on both splits, and the best-effort and $k{=}8$ points sit a few points above it. Each Llama observation is a first-turn instance. The Llama oracle intervals lie below zero, but we treat that as a sample-specific negative-regret result rather than evidence that eviction generally improves models. A 14B bootstrap is omitted because decision-point resampling does not account for correlated turns.

\begin{table}[ht]
\centering
\small
\begin{tabular}{lcccc}
\toprule
\textbf{Model} & \textbf{succ.} & \textbf{oracle} & \textbf{+1 rand} & \textbf{random} \\
\midrule
Qwen2.5-7B base   & 46.4\% & $-0.7$ & $-1.4$ & $+45.0$ \\
Qwen2.5-7B lc     & 40.8\% & $-2.5$ & $-3.3$ & $+38.3$ \\
\midrule
Llama-3.1-8B base & 38.5\% & $-5.0$ & $-2.5$ & $+33.0$ \\
Llama-3.1-8B lc   & 38.5\% & $-7.0$ & $-2.0$ & $+33.0$ \\
\bottomrule
\end{tabular}
\caption{First-turn uncompressed success and hindsight-oracle/control regret (pp) across model families.}
\label{tab:app-crossfamily}
\end{table}

Table~\ref{tab:app-crossfamily} repeats the four conditions on Llama-3.1-8B first turns, next to Qwen2.5-7B first-turn rows evaluated under the same protocol. Oracle and +1 rand eviction stay near or below zero on both families, while same-count random eviction is catastrophic.

\begin{table}[ht]
\centering
\small
\begin{tabular}{lccc}
\toprule
\textbf{Condition} & \textbf{evicted} & \textbf{succ.} & \textbf{regret} \\
\midrule
baseline          & 0.0\%  & 39.7\% & $0.0$ \\
oracle            & 84.5\% & 42.4\% & $-2.7$ \\
oracle +1 rand    & 81.3\% & 41.8\% & $-2.1$ \\
same-count random & 84.9\% & 3.9\%  & $+35.8$ \\
\bottomrule
\end{tabular}
\caption{Qwen2.5-7B results on 330 BFCL \texttt{multi\_turn\_miss\_param} decision points.}
\label{tab:app-miss-param}
\end{table}

Table~\ref{tab:app-miss-param} repeats the causal control on the first 100 BFCL \texttt{multi\_turn\_miss\_param} instances, whose initial requests omit required parameters, producing 330 scored decision points. The evicted column is the mean fraction of prefix tokens removed. The random condition is matched to the oracle by the number of evicted tools, so its token fraction differs slightly because schemas have different lengths.

\subsection{\approach Frontier}

The base-split keep-top-$k$ frontier for both predictors is plotted in the frontier figure of the main paper. For reference, the LLM-select operating point is 4.7 points of regret at 17\% retained and the oracle ceiling is $-1.3$ at about 14\% retained.

\begin{table}[ht]
\centering
\small
\begin{tabular}{c cc@{\hspace{1.4em}} cc}
\toprule
\multirow{2}{*}{\textbf{\boldmath$k$}} & \multicolumn{2}{c}{\textbf{Qwen2.5-14B}} & \multicolumn{2}{c}{\textbf{Llama-3.1-8B}} \\
\cmidrule(r){2-3}\cmidrule(l){4-5}
 & BM25 & LLM & BM25 & LLM \\
\midrule
1  & 27.5 & 24.1 & 16.5 & 20.0 \\
2  & 21.0 & 16.1 & 6.5  & 9.5 \\
3  & 14.8 & 13.0 & 5.0  & 6.5 \\
5  & 9.6  & 9.6  & 3.5  & 4.0 \\
8  & 6.2  & 9.3  & 3.5  & 4.0 \\
12 & 4.4  & 7.8  & 0.0  & 2.0 \\
16 & 1.6  & 7.0  & 1.5  & 2.0 \\
\bottomrule
\end{tabular}
\caption{Base-split keep-top-$k$ regret (pp) across scale and family. Llama uses first-turn decision points.}
\label{tab:app-frontier-xmodel}
\end{table}

Table~\ref{tab:app-frontier-xmodel} repeats the frontier on Qwen2.5-14B and Llama-3.1-8B, whose oracle point estimates are $-1.3$ and $-5.0$. The retained prefix fraction grows with $k$ as on 7B, from about 10\% at $k{=}1$ to about 66\% at $k{=}16$. The ranker trend replicates at a second Qwen scale and on Llama first turns. It is not a full cross-family rollout or CRC evaluation.

\begin{table}[ht]
\centering
\small
\setlength{\tabcolsep}{4pt}
\begin{tabular}{l ccccccc}
\toprule
\textbf{Model} & $k{=}1$ & $2$ & $3$ & $5$ & $8$ & $12$ & $16$ \\
\midrule
Qwen2.5-7B  & 20.7 & 12.8 & 8.8  & 6.1  & 4.9 & 3.3 & 3.0 \\
Qwen2.5-14B & 23.7 & 16.1 & 13.4 & 10.3 & 9.4 & 7.3 & 6.4 \\
\bottomrule
\end{tabular}
\caption{Long-context keep-top-$k$ task-success regret (pp) for the small-model ranking.}
\label{tab:app-frontier-lc}
\end{table}

Table~\ref{tab:app-frontier-lc} gives the same frontier on the long-context split for the small-model ranking at both scales, where the oracle ceiling is $-3.2$ for 7B and $-1.3$ for 14B. The frontier holds there too, so the method is not limited to the schema-heavy base split.

\subsection{Conformal Risk Control}
\label{app:crc}

\begin{table}[ht]
\centering
\small
\setlength{\tabcolsep}{3pt}
\begin{tabular}{llcccc}
\toprule
\textbf{Setting} & $\boldsymbol{\alpha}$ & \textbf{selected $\boldsymbol{k}$} & \textbf{ratio} & \textbf{held-out loss} & \textbf{splits $\boldsymbol{\le\alpha}$} \\
\midrule
7B base & 0.10 & 16 [16, 16]  & 1.47$\times$ & 0.074 & 71\% \\
7B base & 0.20 & 12 [8, 12]   & 2.08$\times$ & 0.154 & 72\% \\
7B lc   & 0.10 & 16 [16, all] & 1.23$\times$ & 0.068 & 62\% \\
7B lc   & 0.20 & 12 [8, 16]   & 1.75$\times$ & 0.167 & 65\% \\
\midrule
14B base & 0.10 & 16 [16, all] & 1.42$\times$ & 0.077 & 69\% \\
14B base & 0.20 & 16 [8, 16]   & 1.87$\times$ & 0.129 & 69\% \\
14B lc   & 0.10 & 16 [16, all] & 1.33$\times$ & 0.072 & 71\% \\
14B lc   & 0.20 & 16 [8, 16]   & 1.66$\times$ & 0.119 & 75\% \\
\bottomrule
\end{tabular}
\caption{Instance-level CRC diagnostics on the BM25 grid under the breakage loss, over 400 instance-disjoint 80/20 splits. Selected $k$ is the median and [10th, 90th percentile] across splits.}
\label{tab:app-conformal}
\end{table}

Table~\ref{tab:app-conformal} expands the main CRC result with the selection range across 400 instance-disjoint 80/20 splits. The mean held-out loss is below $\alpha$ in all eight settings. The ratio column is the mean eviction-only reduction of the selected policies, not a guaranteed minimum, and includes no 4-bit factor. The last column is the fraction of finite test splits whose empirical mean happens to lie below $\alpha$. CRC controls expected loss over calibration data and a fresh instance, so this fraction is a diagnostic rather than a coverage guarantee.

\paragraph{Exchangeable unit and monotone majorant} The eligible sets contain 111/83 instances for 7B base/lc and 112/90 for 14B. Each instance contributes one curve, its mean failure over baseline-correct turns. A raw curve is non-monotone for 17/111, 12/83, 9/112 and 7/90 instances, respectively. We apply the pointwise majorant of the main paper separately to every instance before averaging. The mean increase from this safe closure is 0.010--0.028. This construction supplies the monotone loss required by the original CRC result, which an envelope applied only after empirical averaging would not.

\paragraph{Candidate grids} Qwen-7B uses $k=\{1,2,3,5,8,12,16,\text{all}\}$, and Qwen-14B uses $k=\{3,5,8,16,\text{all}\}$ because only those generations are available. The shared $\alpha=0.10$ result selects $k=16$. However, at $\alpha=0.20$ the missing 14B $k=12$ point can only make its selected policy more conservative, so we do not interpret the 7B--14B compression difference as a scale effect.

\paragraph{Full-data calibration} Calibrating the BM25 grid on all eligible instances selects $k{=}16$ in all four settings, a 1.45--1.51$\times$ eviction-only reduction. At $\alpha=0.10$ the corrected calibration risks are 0.083, 0.097, 0.088 and 0.090 for 7B base/lc and 14B base/lc, and the certified guarantee is $\mathbb{E}[\ell]\le\alpha$.

\paragraph{Breakage-loss certificates} The decode-jitter floor of the breakage loss is measured directly. The instance-level breakage of the hindsight oracle is 0.078 (7B base) and 0.060 (7B lc), so at $\alpha=0.10$ (threshold 0.092 at $n{=}111$) even perfect prediction leaves 0.014 of budget. The small-model full-ranking family (truncated rankings completed by the keep-all endpoint) certifies 3.73$\times$/4.39$\times$ (7B base) and 2.84$\times$/2.96$\times$ (7B lc) at $\alpha=0.15$/$0.20$ while abstaining at $\alpha=0.10$, and 14B lc abstains at every budget.

\paragraph{Llama recalibration} The Llama first-turn set has 77 eligible instances, and 21/77 raw instance curves contain a non-monotone reversal. With the same pointwise closure and keep-all endpoint, full-data calibration at $\alpha=0.10$ selects keep-all. Over 400 80/20 splits, keep-all is selected about 90\% of the time and mean held-out loss is about 0.02. At $\alpha=0.20$, full-data calibration selects $k=8$ with corrected calibration risk 0.179. Over the repeated splits, the median policy is also $k=8$ and mean raw held-out loss is 0.141. The corrected procedure therefore abstains at the tighter budget rather than certify an unsupported point.

\subsection{Attention-Eviction Baselines}
\label{app:attn}

\begin{table}[ht]
\centering
\small
\begin{tabular}{lcc}
\toprule
\textbf{retained} & \textbf{H2O-style} & \textbf{SnapKV-style} \\
\midrule
\multicolumn{3}{l}{\textit{Token-level, direct token budget}} \\
10\% & 34.6 & 33.2 \\
20\% & 25.9 & 21.7 \\
30\% & 24.1 & 15.4 \\
40\% & 17.3 & 10.5 \\
50\% & 12.8 & 8.1 \\
60\% & 11.3 & 7.6 \\
\midrule
\multicolumn{3}{l}{\textit{Per-tool granularity, keep top-$k$ tools}} \\
19\% & 33.2 & 10.2 \\
36\% & 28.8 & 6.3 \\
64\% & 17.5 & 3.9 \\
\bottomrule
\end{tabular}
\caption{Qwen2.5-7B base global attention-score regret (pp) at token and tool granularity.}
\label{tab:app-attention}
\end{table}

Table~\ref{tab:app-attention} gives the token-level H2O-style and SnapKV-style curves and their per-tool-granularity variants. The adaptations aggregate eager prefill attention across layers and heads into one global score, compute the H2O-style score once per decision point, and preserve attention sinks and a recent window in the token-level variants. They are reported on the base split and are not faithful official implementations.

\begin{table}[ht]
\centering
\small
\begin{tabular}{lcc}
\toprule
\textbf{retained} & \textbf{H2O-style} & \textbf{SnapKV-style} \\
\midrule
10\% & 38.0 & 30.5 \\
20\% & 38.0 & 12.0 \\
30\% & 35.0 & 3.5 \\
40\% & 21.0 & $-0.5$ \\
50\% & 9.5  & 1.5 \\
60\% & 3.5  & $-1.0$ \\
\bottomrule
\end{tabular}
\caption{Llama-3.1-8B first-turn global attention-score regret (pp) at each retained prefix fraction.}
\label{tab:app-attention-xmodel}
\end{table}

Table~\ref{tab:app-attention-xmodel} repeats the token-level score baselines on Llama-3.1-8B by spreading the eager attention map across GPUs. The SnapKV-style score approaches baseline only after retaining 30\%--60\% of the prefix, more than \approach uses at its best-effort point. This is a cross-family replication of the score comparison, not a claim against the official method.

\subsection{Always-On Rollout and Memory}

\begin{table}[ht]
\centering
\small
\setlength{\tabcolsep}{3pt}
\begin{tabular}{lccccc}
\toprule
\textbf{policy} & \textbf{evicted} & \textbf{turn succ.} & \textbf{turn regret} & \textbf{ep.\ succ.} & \textbf{calls} \\
\midrule
baseline & 0.0\% & 23.3\% & --- & 6\% & 677 \\
BM25 $k{=}16$ & 33.7\% & 21.5\% & $+1.8$ & 3\% & 669 \\
BM25 $k{=}8$ & 59.7\% & 17.6\% & $+5.8$ & 2\% & 657 \\
\bottomrule
\end{tabular}
\caption{Always-on rollout results over 100 episodes and 330 turns. Regret is computed on unrounded turn success.}
\label{tab:app-e2e}
\end{table}

Table~\ref{tab:app-e2e} details the always-on rollout summarized in the main text. It rolls out 100 base instances from self-generated history with BM25 eviction active at every generation step. Regret stays within a few points of the reference-replay protocol at matched keep sizes, and total emitted calls differ by at most 3\%. No conformal guarantee is claimed for this shifted distribution.

\begin{table}[ht]
\centering
\small
\begin{tabular}{lccc}
\toprule
\textbf{Split} & \textbf{KV full} & \textbf{KV kept} & \textbf{reduction} \\
\midrule
base & 214 MB & 30 MB & 7.1$\times$ \\
lc   & 260 MB & 64 MB & 4.1$\times$ \\
\bottomrule
\end{tabular}
\caption{Measured Qwen2.5-7B prefix KV after physically compacting the hindsight-oracle keep set, 50 decision points per split.}
\label{tab:app-hbm}
\end{table}

Table~\ref{tab:app-hbm} reports the measured Qwen2.5-7B prefix cache after physically compacting the oracle keep set. This validates the memory arithmetic of compacted prefill. The practical predictors and the 4-bit factor are projections.

\subsection{Predictor Details}

\begin{table}[ht]
\centering
\small
\begin{tabular}{llcccc}
\toprule
\textbf{Split} & \textbf{ranker} & \textbf{@1} & \textbf{@3} & \textbf{@5} & \textbf{@8} \\
\midrule
\multirow{2}{*}{base} & BM25 & 26\% & 49\% & 60\% & 70\% \\
     & LLM  & 30\% & 56\% & 63\% & 66\% \\
\midrule
\multirow{2}{*}{lc}   & BM25 & 24\% & 47\% & 57\% & 67\% \\
     & LLM  & 39\% & 57\% & 60\% & 60\% \\
\bottomrule
\end{tabular}
\caption{All-reference-tools recall@$k$ of the BM25 and LLM rankers.}
\label{tab:app-recall}
\end{table}

\paragraph{Predictor recall} Table~\ref{tab:app-recall} reports how often all reference-invoked tools fall inside the top $k$. At $k{=}5$ the rankers reach only about 60\%, and they climb slowly because the missed tools are procedural and implicit ones that the user text never names.

\paragraph{Predictor prompt} The small model sees a system instruction and a user message. The user message lists each tool as its name and the first sentence of its description, followed by the request. The default instruction selects the needed tools. The full-ranking variant used for the predictor frontier requests an order over all tools, but generations can terminate before listing the full catalog. We parse the emitted partial ranking by retaining valid tool names in order and discarding duplicates. Listing~\ref{lst:prompts} shows the two system instructions.

\begin{listing}[t]
\begin{lstlisting}[numbers=none, columns=fullflexible, keepspaces=true, breakindent=0pt, breakautoindent=false]
# default, select the needed tools
You are a tool-selection assistant. Given a user request and a catalog of available tools, output ONLY the names of the tools needed to fulfill the request, as a comma-separated list ordered by importance, most important first. Include every tool required, including implicit setup steps. Output just the tool names.

# full ranking, used for the predictor frontier
You are a tool-ranking assistant. Given a user request and a catalog of available tools, rank ALL of the available tools from most to least likely to be needed. Output every tool name in that order, most likely first. Output just the tool names.
\end{lstlisting}
\caption{Selection and full-ranking instructions for the small-model predictor.}
\label{lst:prompts}
\end{listing}

\subsection{Implementation}
\label{app:impl}

\paragraph{Hyperparameters and seeds} BM25 uses $k_1{=}1.5$, $b{=}0.75$ over lowercased alphanumeric tokens of the name and description of each tool, with the corpus rebuilt per instance from its own tool catalog. The attention-score controls always keep the first 4 tokens (attention sinks) and a recent window. The SnapKV-style score attends from the last 32 query tokens, which are themselves always kept, and the H2O-style score is computed once from the same eager pass. Mixed-precision probes quantize keys per channel and values per token, asymmetric with zero point, with group size 32 on the compressed segment and 128 on the 4-bit base. Random eviction controls are seeded deterministically per decision point by its instance id and turn index, so every condition is exactly reproducible.

\paragraph{Policy simulation and compaction check} We simulate eviction by zeroing the attention mask at the chosen positions. The generation stack re-indexes positions by the cumulative sum of the mask, so masked generation is equivalent to prefilling the compacted prefix without those tokens, and a paired masked-versus-compacted run produces identical outputs. The no-op mask reproduces uncompressed legality and success. For the oracle memory check we build the compacted prefix and read the resulting cache size. We have not implemented mutation of an already-populated cache, practical-predictor compaction, or on-demand reinsertion. The HBM result therefore validates the compaction mechanism, not an end-to-end cache runtime.

\end{document}